%% file: main.tex
\documentclass[]{bytedance_seed}

\usepackage[toc,page,header]{appendix}
\usepackage{minitoc}
\usepackage{amsmath} 
\usepackage{bm} 
\usepackage{booktabs}
\usepackage{graphicx}
\usepackage{makecell}
\usepackage{lipsum}
\usepackage{enumitem}
\usepackage{algorithm}
\usepackage{algpseudocode}
\usepackage{amsmath,amssymb,amsthm}
\usepackage[most]{tcolorbox}
\usepackage{stmaryrd}
\usepackage{todonotes}
\usepackage{amsthm}
\usepackage{parskip} 
\usepackage{wrapfig}
\usepackage{subcaption}
\usepackage{xspace}
\usepackage{caption}
\usepackage{array}
\usepackage{multirow}
\usepackage{pifont}
\usepackage{cleveref}
\usepackage{eucal}
\usepackage{longtable}
\usepackage{amssymb}
\usepackage{CJKutf8}
\newcommand{\method}{\textsc{Seed-X}\xspace}
\usepackage{xcolor} 

\definecolor{Blue}{rgb}{0, 0, 1}
\definecolor{Green}{rgb}{0, 0.5, 0}
\definecolor{MyGray}{gray}{0.9}
\colorlet{MyBlue}{Blue!15}  
\colorlet{MyGreen}{Green!15}

\title{\method: Building Strong Multilingual Translation LLM with 7B Parameters}

\author{ByteDance Seed}

\contribution[]{Full author list in Contributions}

\abstract{
Multilingual translation stands as a challenging task for large language models (LLMs) to handle intricate language patterns and stilted translations that arise in automated translations. 
In this paper, we introduce \textbf{\method}, a family of open-source LLMs comprising instruct and reasoning models, pushing the limits of translation capability with 7B parameter size. 
The base model is pre-trained on a diverse, high-quality dataset encompassing both monolingual and bilingual content across 28 languages, harnessing the full potential of multilingual data.
The instruct model is then finetuned to translate by Chain-of-Thought~(CoT) reasoning and further enhanced through reinforcement learning~(RL) to achieve better generalization across diverse language pairs.
\method achieves performance comparable to leading closed-source models, including Gemini-2.5 and GPT-4o, across 28 languages, and significantly outperforms larger open-source models in both automatic metrics and human evaluations. 
We share the best practices through our optimization process, and make the parameter public available for advancing translation research and applications.

}

\date{\today}
\correspondence{\email{chengshanbo@bytedance.com}}

\checkdata[Project Page]{\url{https://github.com/ByteDance-Seed/Seed-X-7B}}

\begin{document}
\maketitle
\begin{figure}[!h]
    \centering
    \includegraphics[width=1\linewidth]{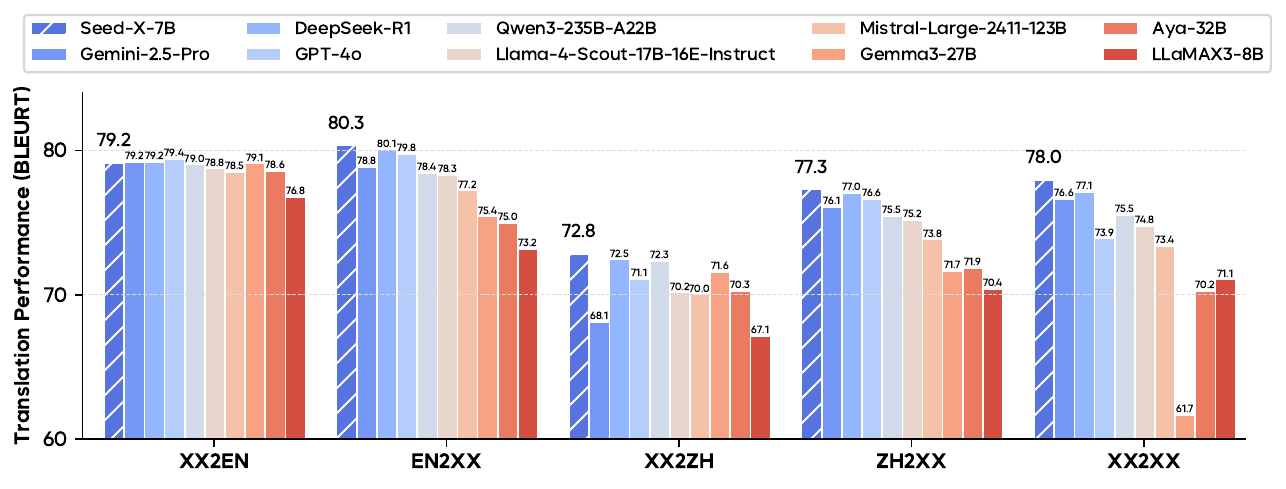}
    \caption{Benchmark performance on Flores-200 of \method and its counterparts.}
    \label{fig:enter-label}
\end{figure}

\input{sections/001-intro}
\input{sections/002-methods}
\input{sections/003-experimets}
\input{sections/004-discussion}

\input{sections/005-related}
\input{sections/006-conclusion}

\clearpage

\bibliographystyle{unsrtnat}
\bibliography{main}

\clearpage

\beginappendix

\input{appendix/language}
\input{appendix/experimental_set}
\input{appendix/ppo_vs_sft}
\input{appendix/human_eval}
\input{appendix/case_study_new}
\newpage
\input{appendix/contributions}
\end{document}

%% file: sections/001-intro.tex

\begin{figure}[htb]
    \centering
    \begin{minipage}{0.48\textwidth}
        \centering
        \includegraphics[width=0.93\linewidth]{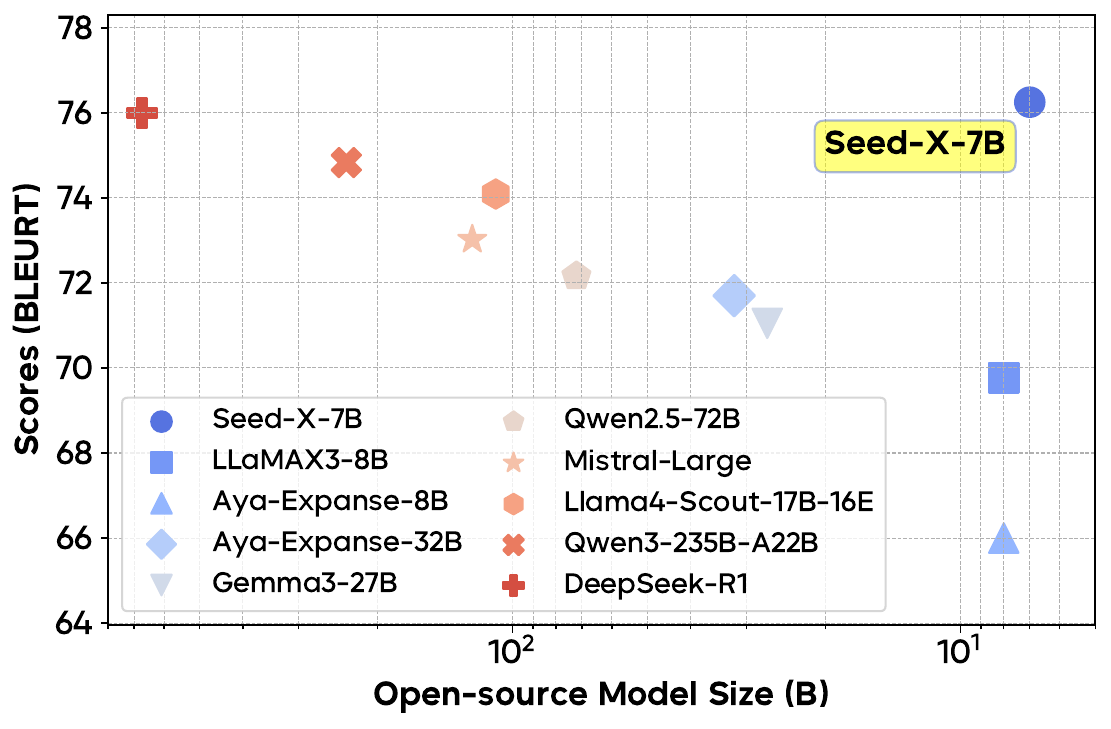}
        \caption{\textbf{Model size} versus average \textbf{multilingual translation performance} for different-scale open-source models. Scores averaged over directions and test sets (see Table~\ref {tab:main_results}).}
        \label{fig:size-bleurt}
    \end{minipage}
    \hfill
    \begin{minipage}{0.48\textwidth}
        \centering
        \includegraphics[width=\textwidth]{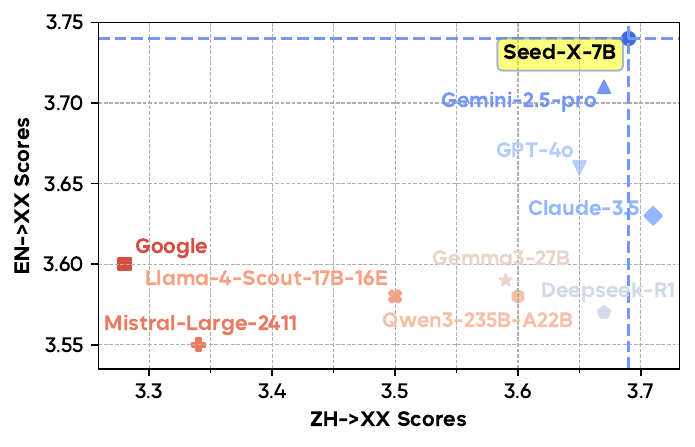}
    \caption{\textbf{Human evaluation} (0-4) of models on \textbf{\method-Challenge} for Chinese/English to 7 languages translation (detailed scores are presented in Appendix~\ref{appendix:human_evaluation}).}\label{fig:humen_eval}
    \end{minipage}
\end{figure}

\section{Introduction}
Machine translation has long been a high-demand practical goal and a significant research challenge pursued over the past few decades~\cite{brown1993mathematics, wu2016google, vaswani2017attention, nllb}. 
Recently, the development of large language models (LLMs) has revolutionized the learning paradigm of machine translation, shifting it towards large-scale learning approaches~\cite{garcia2023unreasonable, zhu2024multilingual, xu2024paradigm, tower_llm_2024}.
The continuous evolution of LLMs has consistently pushed the boundaries of translation quality, with state-of-the-art models like Claude-3.5~\cite{claude35} even surpassing human experts on certain translation directions~\cite{kocmi2024findings}.

However, the challenge has not been completely resolved, such as the translation of idioms, slang, and internet buzzwords~\cite{pang2025salute}.
This requires not only robust language comprehension but also the ability to produce culturally appropriate and natural expressions, going beyond mere word-for-word mapping between languages.
Moreover, a persistent performance gap exists between proprietary and open-source models~\cite{kocmi2024findings}, particularly given the typically limited scale of open-source systems.
The lack of clear methodologies for developing cutting-edge LLM-based translation systems continues to hinder community efforts to deploy and iterate effective translation solutions.

In this paper, we present \method, a comprehensive family of 7B-parameter open-source translation models. The suite encompasses an instruct model and an RL model together with the corresponding reward model. 
The entire training process is designed to address the challenges of cross-lingual learning and enhance both computational efficiency and translation performance. 
First, the models undergo pre-training on a large-scale dataset comprising 28 languages from scratch, incorporating both monolingual and parallel corpora to ensure comprehensive linguistic coverage. 
Subsequently, the instruct model learns to implement chain-of-thought reasoning to facilitate the translation of high-resource languages and finally generalizes to a broader language spectrum by reinforcement learning.

We conduct extensive evaluations on various representative translation benchmarks and compare it with both open-source models and closed-source systems. 
\method demonstrates superior performance compared to translation-specialized models of similar size and Google-Translator. It further outperforms numerous larger-scale LLM systems (Figure \ref{fig:size-bleurt}), including Gemma3-27B~\cite{gemma}, Llama4-Scout-17B-16E~\citep{meta2025llama} and Qwen3-235B-A22B~\citep{yang2025qwen3}. 

Human evaluation in Figure~\ref{fig:humen_eval} shows that its performance matches or even surpasses that of cutting-edge models including GPT-4o~\citep{GPT-4o}, Claude-3.5~\citep{claude35}, DeepSeek-R1~\citep{deepseekai2025deepseekr1}, and Gemini-2.5-Pro~\citep{gemini25}. 
Overall, we set a higher-level baseline for machine translation research based on open-source 7B models by providing:
\begin{itemize}
    \item \textbf{A family of \method models} including the instruct model, RL model, and reward model.
    \item \textbf{A diverse and challenging translation testset} comprising English/Chinese$\Rightarrow$X idioms, proverbs, poetry, internet slang, and abbreviations across various domains, accompanied by annotated translation key points.
    \item \textbf{A detailed training recipe} for developing translation-oriented LLMs, along with insights into multilingual knowledge transfer mechanisms.
\end{itemize}

%% file: sections/002-methods.tex
\section{Pre-training}\label{s:pretraining}

This section describes the construction of our pre-training dataset, which incorporates both monolingual and bilingual corpora to facilitate effective multilingual modeling and cross-language semantic alignment.

\subsection{Monolingual Data}\label{sec:mono_data}

\input{tables/pretrain_mono_data}
The monolingual data comprises a diverse blend of languages with a total of 6 trillion tokens (see Table~\ref{tab:pretrain_mono_data_lang}). Our data pre-processing pipeline is designed to optimize both data quality and linguistic diversity across language corpora while minimizing redundancy. We collect extensive monolingual data from various sources, implement document tagging, and balance the topic distribution across the corpus~\citep{lu2023instag}. To maintain high data quality, we develop a document-level quality assessment model that categorizes documents into three tiers: high, medium, and low quality. High-quality documents are retained in their original form, medium-quality documents are enhanced through LLM-based paraphrasing, and low-quality content is excluded entirely. This quality control process is implemented iteratively to ensure optimal results.

Besides, we do not spend too much effort on mathematics and code data, allocating the entire data share to multilingual data. We deliberately exclude STEM, coding, and reasoning-focused data, which consume training capacity without contributing significantly to translation quality. 
Thus, we can expect that the base model's capabilities in the field of code will be limited. 

\subsection{Bilingual Data} \label{sec:parallel_data}
Bilingual data plays a crucial role in semantic alignment and can significantly improve translation capabilities. 
We build the multilingual bilingual data in a progressively weak-to-strong fashion using the boosting \method model. 

\begin{itemize}
    \item \textbf{Collecting Seed Bilingual Data.} We create an initial seed bilingual dataset of 200B tokens from publicly available web data. The filtering process employs strict rules based on language identification confidence scores and bilingual alignment metrics from word alignment tools.
    \item \textbf{Building Early Translation Model.} We used the seed bilingual dataset to train an initial translation model, based on either an open-source model or our boosting pretraining \method model. This early model is then fine-tuned for data augmentation, filtering, and rewriting.
    \item \textbf{Augmenting and Refining Bilingual Data.} To further expand the data volume and improve the data quality, we utilize the temporary translation model in the previous phase to construct a new version of bilingual data: (1) directly translating the existing monolingual data to different languages to construct pseudo parallel data, (2) rewriting and paraphrasing the first-round translation, and (3) evaluating and filtering the poor-quality pairs.
    \item \textbf{Repeating the Above Process.} Each iteration of this process progressively enhances the quality of the bilingual data, which is then used to train subsequent model versions. Concurrently, we replace the early translation model in step 2 with more advanced variants to generate successive versions of higher-quality parallel data. Thus, the data quality gradually improves as increasingly sophisticated models are employed.
\end{itemize}

\subsection{Pre-training Stage}\label{sec:training_Stage}

Given the monolingual and bilingual data introduced in Section~\ref{sec:mono_data} and \ref{sec:parallel_data}, the next question is how to organize these datasets, which differ greatly in the volume and distribution of data between languages. Our core strategy focuses on first acquiring fundamental knowledge represented across major languages, then transferring this knowledge to develop multilingual capabilities. The training process follows three key stages:
\begin{itemize}
    \item \textbf{S1: General Stage.} We train the model on large-scale monolingual data from predominant languages, as detailed in Table~\ref{tab:pretrain_mono_data_lang}. The primary languages of the model are Chinese and English.
    \item \textbf{S2: Multilingual-dominant Stage.} We gradually increase the proportion of multilingual data, incorporating both monolingual data from underrepresented languages and bilingual corpora.
    \item \textbf{S3: Parallel-Only Stage.} We fine-tune the model using high-quality bilingual data that has undergone multiple rounds of rewriting and filtering to achieve optimal performance. Specifically, the parallel data is simply combined with language tags, without incorporating any additional instructions or prompts.
\end{itemize}

We build our 7B model based on the Mistral-7B architecture~\citep{jiang2023mistral} and expand the vocabulary size to 65,269 tokens.
To achieve optimal model performance, we develop scaling laws to predict optimal hyper-parameters, such as learning rate scheduler and batch size, specifically tailored for our three pre-training stages. 
More details about the comprehensive experimental settings and configurations can be found in Appendix~\ref{s:Experimental_Settings}.

\subsection{Pre-training Evaluation}
\input{tables/pretrain_eval}
We evaluate the multilingual capabilities of \method after S2, with a focus on knowledge-based tasks rather than mathematical or logical reasoning assessments. Our evaluation framework encompasses two key benchmarks: (1) regional knowledge assessment using INCLUDE~\citep{romanou2024include}, general knowledge evaluated by MMMLU~\citep{hendrycks2020measuring}, and (2) comprehension tasks evaluated through XCOPA~\citep{ponti2020xcopa}, XWinograd~\citep{muennighoff1786crosslingual}, XStoryCloze~\citep{lin2022few}, and PAWS-X~\citep{yang2019paws}. For MMMLU benchmarks, we sample only 10\% of the original data to improve evaluation efficiency. We only report the average scores of languages we support. It's worth noting that S3 employs training data with a particular format that guides \method towards becoming a specialized model. This specialization may have implications for the model's general capabilities, as discussed further in Section~\ref{s:discussion}. 

As shown in Table~\ref{tab:pretrain_eval}, \method demonstrates strong performance in multilingual comprehension tasks, which serves as a fundamental foundation for translation capabilities. \method exhibits strong performance in cultural and region-specific knowledge domains, yet demonstrates relatively lower proficiency in mathematical reasoning tasks within MMMLU.

\section{Post-training}\label{s:post_training}
In the post-training, we aim to equip the base model with robust translation capabilities through supervised fine-tuning (SFT) and preference learning. The central challenge lies in optimizing performance for high-resource languages using human-annotated data while ensuring effective generalization to low-resource languages.

\subsection{Supervised Fine-tuning}\label{ss:sft}

\input{tables/prompt_templates}

We curate our instruction-tuning datasets to include 236K instances to only support translation without general requirements. The dataset draws from two primary sources: the public FLORES devset~\cite{goyal2022flores} and manually annotated translation pairs encompassing both general domains and specific business scenarios. To ensure data quality, we employ G-DIG~\cite{pan2024g} to eliminate problematic examples that contribute most significantly to translation errors. Additionally, we implement rejection sampling~\cite{yuan2023scaling} to further refine the instruction data, with particular emphasis on low-resource languages where manual annotations are scarce.

\textbf{Chain-of-thought Data.} As aforementioned, translation is a complex task that requires reasoning capabilities to comprehend both semantic nuances and cultural context of the source text. To capture this reasoning process, we engage professional linguists to document their thought process when handling challenging translation cases. 
Volunteers are required to annotate the following: (1) the overall sentence meaning, (2) interpretations and translations of challenging linguistic elements (including slang, internet buzzwords, and poetic expressions), (3) notable target language expressions and usage conventions, and common potential translation pitfalls and errors. We design specific CoT prompts to lead the model to translate with the reasoning process as seen in Table~\ref{tab:prompt_templates}.

\subsection{MT-oriented Preference Learning}\label{ss:rlhf}
Large-scale RL has demonstrated effectiveness in enhancing reasoning capabilities for code and mathematical tasks, where outputs follow structured formats~\citep{deepseekai2025deepseekr1}. However, MT presents a unique challenge as its outputs are semantically diverse and resist evaluation through explicit rule-based assessment. For this, we define two distinct types of rewards for different scenarios:

\begin{itemize}
    \item \textbf{Human Preference-based Reward.} Leveraging human preference data, we train a reward model based on \method-base to assign scalar scores to candidate translations. Given the substantial challenges in annotation, including the availability of qualified annotators and associated costs, we focus on preference data collection on high-resource language pairs with approximately 20k pairs.
    \item \textbf{Dual-based Reward without Reference Answer.} When the preference data are unavailable, we follow DuPO~\citep{she2025dupoenablingreliablellm} to employ the dual-based reward to estimate the quality of translation without the need for annotated preference data or reference translations. Specifically, we perform $\rm{A}\Rightarrow{B}\Rightarrow\widetilde{A}$ and measure the similarity between the $\rm{A}$ and $\rm{\widetilde{A}}$ to score the reward of $\rm{A}\Rightarrow{B}$.
\end{itemize}

It is noted that we avoid using automatic evaluation metrics (e.g., BLEU~\citep{papineni2002bleu}, BLEURT~\citep{sellam2020bleurt} or COMET~\cite{xcomet,guerreiro2023xcomet,rei2020comet}) as rewards. These metrics partially reflect translation quality, but are difficult to entirely equivalent to human judgment, especially in assessing subtle linguistic features and contextual appropriateness.

\textbf{RL Algorithm.} We optimize the model parameters using the Proximal Policy Optimization (PPO) algorithm~\citep{schulman2017proximal}. We employ a large batch size combined with multiple rollouts per query for training efficiency and initialize the critic model with the reward model to ensure training stability.

%% file: tables/pretrain_mono_data.tex
\begin{wraptable}{r}{0.35\textwidth}
    \centering
    \small
    \begin{tabular}{ccc}
            \toprule
            \textbf{Language} & \textbf{Prop. (\%)} & \textbf{Tokens (T)} \\
            \midrule
            English & 12.94 & 0.78\\
            Chinese & 10.54 & 0.63\\
            Russian & 8.32 & 0.49\\
            French & 7.36 & 0.44\\
            Spanish & 7.04 & 0.42\\
            German & 6.39 & 0.38\\
            \bottomrule
        \end{tabular}
        \caption{The statistics on high-resource languages in monolingual data.}
        \label{tab:pretrain_mono_data_lang}
\end{wraptable}

%% file: tables/pretrain_eval.tex
\begin{table*}[t]
\centering
\small
\begin{tabular}{lcccccccl}
\toprule

\multirow{2}{*}{\textbf{Model}} & \multicolumn{2}{c}{\textbf{Exam}} & \multicolumn{4}{c}{\textbf{Understanding}} & \multirow{2}{*}{\textbf{Avg.}} \\
  \cmidrule(lr){2-3} \cmidrule(lr){4-7} 
 & \textbf{INCLUDE} & \textbf{MMMLU} &\textbf{XCOPA} & \textbf{XWinograd} & \textbf{XStoryCloze} & \textbf{PAWS-X} & \\  \midrule
Llama3-8B~\citep{Dubey2024TheL3} & 55.27 & 50.60 & 72.03 & 80.25& 65.45 & 60.85 & 64.08\\
Qwen2.5-7B~\citep{Yang2024Qwen25TR} & 61.29 & \textbf{59.59} & 72.73& 79.94& 65.76 & 62.88 & 67.03\\
\method & \textbf{61.93} & 53.25 & \textbf{77.37} & \textbf{82.48} & \textbf{69.72} & \textbf{63.83} & \textbf{68.10}\\
 \bottomrule
\end{tabular}
\caption{Performances of pretrained model on multilingual benchmarks.} \label{tab:pretrain_eval}
\end{table*}

%% file: tables/prompt_templates.tex
\begin{table*}[t]
\small
\centering
\begin{tabular}{lc}
\toprule
\textbf{Prompts}                                                                                                                                             & \textbf{Types}       \\ \midrule
Translate the following text from \texttt{<src>} to \texttt{<trg>}:\underline{\hspace{1em}} & \multirow{4}{*}{Standard}          \\
What does this sentence mean in \texttt{<trg>} from \texttt{<src>}:\underline{\hspace{1em}}  &         \\
How do you translate this sentence into \texttt{<trg>} from \texttt{<src>}:\underline{\hspace{1em}}    &           \\
Translate the following text to \texttt{<trg>}:\underline{\hspace{1em}}        & \\ \midrule
Translate the following \texttt{<src>} sentence into \texttt{<trg>} and explain it in detail:\underline{\hspace{1em}}  & \multirow{3}{*}{CoT} \\
First translate the \texttt{<src>} text into \texttt{<trg>} and then give the explanation:\underline{\hspace{1em}}  & \\
Translate the following sentence into \texttt{<trg>} and try to explain this translation. The input is:\underline{\hspace{1em}} & \\ \bottomrule
\end{tabular}
\caption{A few examples of prompt templates.}\label{tab:prompt_templates}
\end{table*}

%% file: sections/003-experimets.tex
\section{Translation Evaluation}\label{s:experiments}

\subsection{Settings}
\textbf{Benchmarks.} To comprehensively evaluate the multilingual translation capabilities, we conducted extensive experiments using the following test sets: 
\begin{itemize}
    \item \textbf{FLORES and WMT Testsets.} From FLORES-200~\citep{nllb}, we selected 756 translation pairs across 28 languages (listed in Appendix~\ref{s:support_languages}), which were systematically categorized into five groups: English$\Rightarrow$XX, XX$\Rightarrow$English, Chinese$\Rightarrow$XX, XX$\Rightarrow$Chinese, and XX$\Rightarrow$XX translations. Additionally, we incorporated dev sets from WMT-25\footnote{https://www2.statmt.org/wmt25/}, which include English-to-X translations spanning 25 target languages. 
    \item \textbf{\method-Challenge Testset.} Unlike existing datasets that mainly focus on simple news translation, we aim to further challenge and evaluate the boundaries of LLMs' translation capabilities. To this end, we develop a challenging testset containing diverse linguistic elements from internet slang and classical literature to idioms and proper nouns. The dataset spans multiple domains (healthcare, science, tourism, education, e-commerce, entertainment, etc.) and features both formal and colloquial styles. Each entry contains source texts in Chinese or English, with expert annotations highlighting potential translation difficulties. We test translation from Chinese/English to seven languages: Spanish, German, French, Russian, Arabic, Portuguese and Italian.
\end{itemize}

\textbf{Evaluation.} We evaluate translations through two approaches: automatic metrics and human assessment. For automatic evaluation, we employ neural-based metrics XCOMET-XL~\citep{xcomet} and BLEURT~\citep{sellam2020bleurt}, which generally correlate well with human judgements but may be unreliable in detecting certain translation phenomena. Therefore, we conduct human evaluation where multilingual experts score translations on a 0-4 scale, focusing on previously annotated error-prone points and considering accuracy, fluency, and idiomaticity.

\input{tables/main_results}

\subsection{Main Results}

Table~\ref{tab:main_results} demonstrates the competitive performance of our \method on FLORES-200 and WMT-25 benchmarks. For comparison, we categorize baseline models into \textbf{three tiers}: \textbf{(1) ultra-large models} (e.g., Gemini-2.5, Claude-3.5, GPT-4o and so on), \textbf{ (2) medium to small-sized models} such as Gemma-3-27B and Aya-32B, and \textbf{ (3) translation-specialized models} like LLaMAX3-8B and TowerInstruct-13B. It is worth noting that we only evaluate the highest-performing variant from each model series for comparison. Google Translator is the only exception due to its undisclosed model architecture and serving infrastructure.

Super large models dominate the top performance rankings in translation quality, maintaining a significant margin over other models. Their superiority is particularly pronounced in low-resource translation pairs, extending beyond the Chinese-English centered directions. The performance hierarchy among the three tiers is: \textbf{Tier 1 $\gg$ Tier 2 $\approx$ Tier 3}. It also shows that previous specialized translation models, despite their task-specific design, remain constrained by their relatively smaller model sizes, leading to suboptimal performance.

According to automatic evaluation metrics, \method, despite its 7B parameter scale, performs comparably to Tier 1 models and substantially outperforms Tier 2 and Tier 3 models. This shifts the performance hierarchy to \textbf{Tier 1 $\approx$ \method $\gg$ Tier 2~$\approx$ Tier 3}, demonstrating a significant advancement in translation model capabilities.

\textbf{Human Evaluation.} 
Figure~\ref{fig:humen_eval} further demonstrates the effectiveness of \method across 7 widely-used translation language pairs, where professional linguists evaluated translations between Chinese/English and Spanish/German/French/Russian/Portuguese/Italian/Arabic. \method demonstrates superior performance in two aspects: it leads EN$\Rightarrow$XX evaluations, outperforming even ultra-large models, while achieving the second-best results in ZH$\Rightarrow$XX, significantly ahead of mid-tier competitors. Although closed-source large language models typically outperform open-source ones, Seed-X has successfully challenged this dominance. Notably, while Google's system achieves high scores in automatic metrics, it performs relatively poorly in human evaluations~(showcased in Appendix~\ref{s:case_study}). This discrepancy clearly illustrates the limitations of automatic evaluation metrics and validates our motivation to avoid using them directly as rewards in RL.

\subsection{Ablation}

\input{tables/pretrain_ablation_4.3.1}

\textbf{The Effect of Quality and Diversity of Monolingual Data.} This section explores our monolingual data cleaning strategy through a toy experiment on a 1.3B-parameter model, evaluated using two methods: perplexity-based evaluation~\cite{bi2024deepseek} on the translated ARC dataset~\cite{allenai:arc} (``ARC-mm'') and generation-based evaluation on the translation dataset with 5-shot as prompts (``Trans.''). As shown in Table~\ref{tab:mono_quality}, both the filtering and paraphrasing operations significantly reduce the training loss and improve the model performance on the multilingual ARC datasets, which also greatly enhance the LLMs' cross-lingual generalization ability. To measure diversity, we quantify it using the number of unique tags, as shown in Table~\ref{tab:mono_diversity}. The model's reasoning and translation abilities significantly decline as the number of unique tags decreases, highlighting the importance of maintaining diversity in the pre-training dataset.

\begin{figure*}[t]
     \centering
     \begin{subfigure}[b]{0.32\textwidth}
         \centering
         \includegraphics[width=\textwidth]{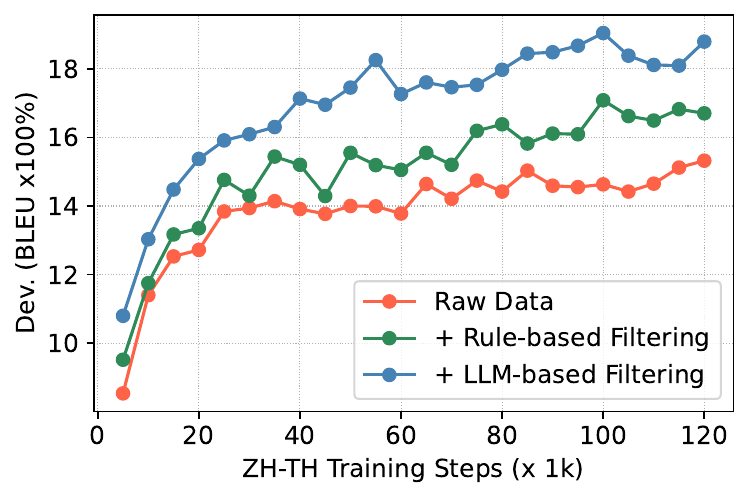}
         \caption{Enc-Dec Model}
         \label{fig:ablation_parallel_enc_enth}
     \end{subfigure}
     \hfill
     \begin{subfigure}[b]{0.32\textwidth}
         \centering
         \includegraphics[width=\textwidth]{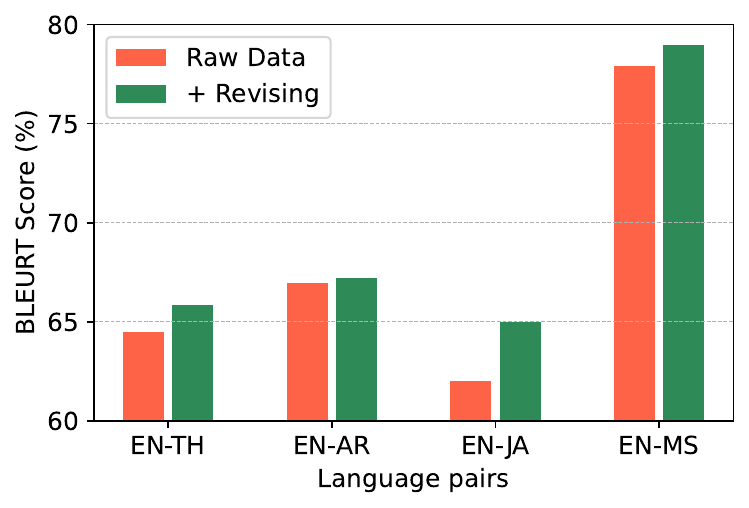}
         \caption{Enc-Dec Model}
         \label{fig:ablation_parallel_enc}
     \end{subfigure}
     \hfill
     \begin{subfigure}[b]{0.33\textwidth}
         \centering
         \includegraphics[width=\textwidth]{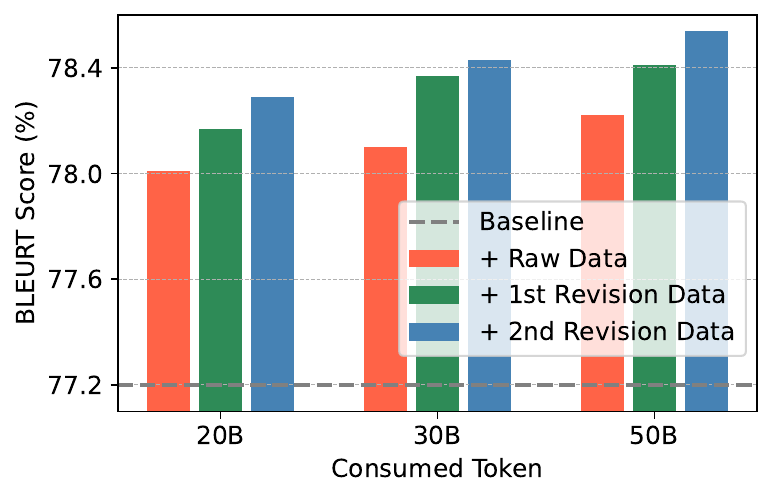}
         \caption{LLM}
         \label{fig:ablation_parallel}
     \end{subfigure}
        \label{lm_ability}
        \caption{Performance gains of revising and filtering parallel data on encoder-decoder models and LLMs.}
\end{figure*}

\textbf{The Effect of Filtering and Revising Parallel Data.} Parallel corpora serve as the most direct supervision signals for translation tasks. Therefore, we design several rounds of filtering and rewriting to progressively improve the quality of parallel corpora. The base model of both filter and rewriter is updated each round to better detect and correct translation errors. To evaluate our approach, we first experiment with Transformer models (13-layer encoder, 3-layer decoder) for each language pair. Figure~\ref{fig:ablation_parallel_enc_enth} shows our LLM-based filtering outperforms rule-based methods in identifying noisy pairs. Figure~\ref{fig:ablation_parallel_enc} demonstrates that our revision method further improves parallel data quality. To assess the impact on LLM performance, we train models on different scales of revised data. Results in Figure~\ref{fig:ablation_parallel} confirm the effectiveness of this multi-round strategy in improving LLM translation performance.


\input{tables/sft_effect}
\textbf{The Effect of Prompt Design.} As a multitasking process, multilingual translation relies heavily on effective prompt design to enhance the model's generalization capabilities. Our findings, as shown in Table~\ref{tab:sft_effect}, demonstrate that omitting the source language in prompts does not impact performance, while making the model more robust when handling mixed-language inputs. Besides, improving the diversity of prompts (as shown in Table~\ref{tab:prompt_templates}) and introducing CoT both greatly enhance the translation quality.

\label{ss:delimiters}

\textbf{The Effect of Delimiters.} Motivated by \cite{wu2021language,liu2024large}, we use language tags \texttt{<LANG\_CODE>} (\texttt{<EN>}, \texttt{<ZH>}, \texttt{<JA>}, etc.) as prefixes and concatenate bilingual data for continued training. Table~\ref{tab:delimiter_results} demonstrates that the incorporation of language tags, similar to language names, in organizing bilingual data can significantly enhance translation quality when compared to other delimiters. Such language-aware delimiters contains much more information than \texttt{<SEP>}, explicitly helping the model distinguish between languages. However, combining bilingual data according to the natural language of the prompt (denoted as \texttt{<NL>}) is formally closer to monolingual data, it introduces a large amount of low-information prompt text. 

%% file: tables/main_results.tex
\begin{table*}[!t]
\centering
\small
\tabcolsep 2pt
\begin{tabular}{p{4.5cm}ccccccccc}
\toprule
\multirow{2}{*}{\textbf{Models}} &  
  \multirow{2}{*}{\textbf{Metrics}} &
  \multicolumn5{c}{\textbf{FLORES-200}} &
  \multicolumn1{c}{\textbf{WMT-25}} &
  \multirow{2}{*}{\textbf{Avg.}} \\ \cmidrule(lr){3-7}\cmidrule(lr){8-8}
                               &    & \textbf{XX$\Rightarrow$EN} & \textbf{EN$\Rightarrow$XX} & \textbf{XX$\Rightarrow$ZH} & \textbf{ZH$\Rightarrow$XX} & \textbf{XX$\Rightarrow$XX} &
                               \textbf{EN$\Rightarrow$XX} &  & \\ \midrule
\multirow{2}{*}{\colorbox{MyGray}{\makebox[4.2cm][l]{TowerInstruct-13B$^\dagger$~\citep{tower_llm_2024}}}} & BLEURT& 76.18 & 68.58    & 66.08 & 51.03 & 56.38 & 56.35 & 62.43 \\
                                                       & COMET  & 93.15 & 79.65    & 78.81 & 83.44 & 81.20 & 66.03 & 80.38 \\\midrule
\multirow{2}{*}{\colorbox{MyGray}{\makebox[4.2cm][l]{LLaMAX3-8B$^\dagger$~\citep{lu2024llamax}}}}       & BLEURT & 76.75 & 73.19    & 67.12 & 70.43 & 71.06 & 60.07 & 69.77 \\
                                                       & COMET  & 95.13 & 86.22    & 78.09 & 83.85 & 83.42 & 70.89 & 82.93 \\\midrule
\multirow{2}{*}{\colorbox{MyGray}{\makebox[4.2cm][l]{Google-Translator}}}                     & BLEURT & 80.13 & 81.07    & 73.81 & 74.84 & 76.93 & 71.07 & 76.31 \\
                                                       & COMET  & 96.98 & 94.76    & 85.67 & 89.64 & 89.96 & 84.12 & 90.19 \\\midrule

\multirow{2}{*}{\colorbox{MyGreen}{\makebox[4.2cm][l]{InternLM3-8B-Instruct$^\dagger$~\citep{cai2024internlm2}}}}
& BLEURT & 70.37 & 56.91    & 62.00 & 50.64 & 40.30 & 47.54 & 54.63 \\
                                                       & COMET  & 85.40 & 63.21    & 71.74 & 59.77 & 58.75 & 53.56 & 65.41 \\\midrule
\multirow{2}{*}{\colorbox{MyGreen}{\makebox[4.2cm][l]{Aya-32B$^\dagger$~\citep{dang2024aya}}}}           & BLEURT & 78.60 & 74.95    & 70.29 & 71.86 & 70.25 & 64.25 & 71.70 \\
                                                       & COMET  & 96.04 & 87.36    & 82.34 & 84.49 & 82.51 & 75.37 & 84.69 \\\midrule
\multirow{2}{*}{\colorbox{MyGreen}{\makebox[4.2cm][l]{Gemma3-27B$^\dagger$~\cite{gemma}}}}               & BLEURT & 79.09 & 75.42    & 71.59 & 71.67 & 61.66 & 66.86 & 71.05 \\
                                                       & COMET  & 96.67 & 92.71    & 84.00 & 89.30 & 85.11 & 81.90 & 88.28 \\\midrule
\multirow{2}{*}{\colorbox{MyGreen}{\makebox[4.2cm][l]{Qwen-2.5-72B-Instruct$^\dagger$~\citep{Yang2024Qwen25TR}}}} & BLEURT & 78.79 & 75.15    & 71.14 & 70.07 & 72.59 & 65.23 & 72.16 \\
                                                       & COMET  & 96.64 & 88.65    & 83.79 & 84.84 & 84.72 & 76.86 & 85.92 \\\midrule
\multirow{2}{*}{\colorbox{MyGreen}{\makebox[4.2cm][l]{Mistral-Large-2411-123B$^\dagger$~\citep{Mistral}}}} & BLEURT & 78.51 & 77.22    & 70.02 & 73.83 & 73.40 & 65.16 & 73.02 \\
                                                       & COMET  & 96.28 & 91.18    & 82.23 & 88.02 & 86.31 & 78.11 & 87.02 \\\midrule
\multirow{2}{*}{\colorbox{MyGreen}{\makebox[4.2cm][l]{Llama-4-Scout-17B-16E$^\dagger$ \citep{meta2025llama}}}} & BLEURT & 78.75 & 78.31    & 70.16 & 75.18 & 74.77 & 67.37 & 74.09 \\
                                                       & COMET  & 96.25 & 92.45    & 83.11 & 89.32 & 87.80 & 80.44 & 88.23 \\\midrule

\multirow{2}{*}{\colorbox{MyBlue}{\makebox[4.2cm][l]{Qwen3-235B-A22B$^\dagger$~\citep{yang2025qwen3}}}}
& BLEURT & 79.03 & 78.43    & 72.32 & 75.47 & 75.55 & 68.19 & 74.83 \\
                                                       & COMET  & 96.75 & 92.66    & 85.38 & 89.55 & 89.00 & 81.45 & 89.13 \\\midrule
\multirow{2}{*}{\colorbox{MyBlue}{\makebox[4.2cm][l]{Gemini-2.5-Pro~\citep{gemini25}}}} & BLEURT & 79.18 & 78.83    & 68.13 & 76.08 & 76.63 & 69.23 & 74.68 \\
                                                       & COMET  & 96.52 & 93.16    & 79.79 & 90.59 & 89.80 & 83.17 & 88.84 \\\midrule
\multirow{2}{*}{\colorbox{MyBlue}{\makebox[4.2cm][l]{GPT-4o~\citep{GPT-4o}}}} & BLEURT & 79.40 & 79.78    & 71.11 & 76.64 & 73.90 & 70.08 & 75.15 \\
                                                       & COMET  & 97.08 & 94.20    & 83.04 & 91.56 & 90.08 & 83.47 & 89.90 \\\midrule
\multirow{2}{*}{\colorbox{MyBlue}{\makebox[4.2cm][l]{DeepSeek-R1$^\dagger$~\citep{deepseekai2025deepseekr1}}}} & BLEURT&79.17& 80.06   & 72.45 & 77.05 & 77.14 & 71.06 & 76.16 \\
                                                      & COMET  & 96.89 & 94.53    & 85.27 & 91.43 & 90.77 & 84.82 & 90.62 \\\midrule
\multirow{2}{*}{\colorbox{MyBlue}{\makebox[4.2cm][l]{Claude-3.5-Sonnet~\citep{claude35}}}}
& BLEURT & 79.66 & 80.53    & 72.65 & 77.76 & 78.14 & 71.06 & 76.63 \\
                                                      & COMET  & 97.17 & 94.88    & 85.47 & 92.36 & 91.71 & 84.82 & 91.07 \\\midrule
\multicolumn{9}{c}{\textcolor{blue}{\textbf{Our Systems}}} \\ \midrule

\multirow{2}{*}{\method-Instruct$^\dagger$}                      & BLEURT & 77.77 & 79.46    & 72.16 & 75.51 & 76.38 & 68.90 & 75.03 \\
                                                       & COMET  & 95.52 & 93.51    & 85.54 & 89.18 & 88.86 & 81.69 & 89.05 \\\midrule
\multirow{2}{*}{\method-DuPO$^\dagger$}                          & BLEURT & 78.57 & 80.13    & 72.57 & 77.05 & 77.35 & 69.32 & 75.83 \\
                                                       & COMET  & 96.55 & 94.03    & 85.76 & 91.02 & 90.43 & 81.99 & 89.96 \\ \midrule
\multirow{2}{*}{\method-PPO$^\dagger$}                           & BLEURT & 79.15 & 80.34    & 72.83 & 77.32 & 77.98 & 69.88 & 76.25 \\
                                                       & COMET  & 96.98 & 94.60    & 87.24 & 91.80 & 91.34 & 82.89 & 90.81 \\ 
\bottomrule
\end{tabular}
\caption{Performances of different models on FLORES-200 and WMT-25. We report the English-centric (EN$\Rightarrow$XX and XX$\Rightarrow$EN), Chinese-centric (ZH$\Rightarrow$XX and XX$\Rightarrow$ZH) and the overall XX$\Rightarrow$XX performances of \method and prominent existing systems. Systems with open-source weights are marked with $\dagger$. Baseline models are categorized into three groups: (1) \colorbox{MyBlue}{\textbf{ultra-large general models}}, (2) \colorbox{MyGreen}{\textbf{medium to small-sized general models}}, and (3) \colorbox{MyGray}{\textbf{translation-specialized models}}.}
\label{tab:main_results}
\end{table*}

%% file: tables/pretrain_ablation_4.3.1.tex
\begin{table*}[htbp]
    \centering
    \small
    \tabcolsep 3pt
    \begin{subtable}[h]{0.49\linewidth}
            \centering
        \begin{tabular}{lccc}
            \toprule
        \textbf{Model} & \textbf{Loss} & \textbf{Arc-mm} & \textbf{Trans.}  \\
     \midrule
        \textsc{Raw Data} & 3.23 & 25.17 & 6.97 \\
        $+$ \textsc{Filter} & 2.93 & 30.18 & 8.19 \\
         $+$ \textsc{Paraphrasing} & \textbf{2.81} & \textbf{32.98} & \textbf{9.41} \\
         \bottomrule   
        \end{tabular}
        \caption{Impact of quality improvements in pre-training data.}
        \label{tab:mono_quality}
    \end{subtable}
    \begin{subtable}[h]{0.49\linewidth}
    \centering
        \begin{tabular}{lcccc}
    	\toprule
            \textbf{Model} & \textbf{Loss} & \textbf{Arc-mm} & \textbf{Trans.} & \textbf{\# Tag (M)}  \\
            \midrule
            \textsc{Baseline} & 2.81 & \textbf{32.98} & \textbf{9.41} & \textbf{1.55} \\
            \textsc{$2/3$ Diversity} & 2.73 & 29.71 & 8.55 & 1.04 \\
            \textsc{$1/3$ Diversity} & \textbf{2.51} & 26.51 & 7.67 & 0.51 \\
            \bottomrule      
        \end{tabular}
        \caption{Effect of diversity in pre-training data.}
        \label{tab:mono_diversity}
    \end{subtable}
    \caption{Performance improvements achieved through monolingual data quality enhancement and diversity selection strategies. We observe significant gains in translation ability as the quality of monolingual data improves.}
\end{table*}

%% file: tables/sft_effect.tex
\begin{table}[t]
    \centering
    \small
    \begin{minipage}{0.45\textwidth}
        \centering
        \begin{tabular}{lcc}
            \toprule
            \textbf{SFT} & \textbf{EN$\Rightarrow$XX} & \textbf{XX$\Rightarrow$EN} \\
             \midrule
              \textsc{Baseline}  & 75.84  & 75.66           \\ \midrule
              \quad \textit{w/} \textsc{No [\textit{src}] Lang.}  & 75.82 & 75.89   \\
              \quad \textit{w/} \textsc{CoT Prompts}  & 77.48     &77.58       \\
              \quad \textit{w/} \textsc{Diversed Prompts} & 77.17     & 77.5       \\
              \quad \textit{w/} \textsc{All} & \textbf{77.88}   & \textbf{77.74}     \\ \bottomrule
        \end{tabular}
        \vspace{5pt} 
        \caption{Ablation of different prompts in SFT.}
        \label{tab:sft_effect}
    \end{minipage}
    \hspace{0.7cm}  
    \begin{minipage}{0.45\textwidth}
        \centering
        \begin{tabular}{lcc}
            \toprule
            \textbf{Delimiter} & \textbf{EN$\Rightarrow$XX} & \textbf{XX$\Rightarrow$EN} \\  \midrule
            \textsc{\textless{}SEP\textgreater{}}   & 53.39 & 74.97   \\
            \textsc{\textless{}NL\textgreater{}}       & 74.81 & 75.47         \\
            \textsc{\textless{}Language\textgreater{}}     & 77.66    & 77.75   \\
            \textsc{\textless{}Lang Code\textgreater{}}   & \textbf{78.31}   & \textbf{77.86}     \\ \bottomrule
        \end{tabular}
        \caption{Comparison of different delimiters used for concatenating parallel data. \texttt{<NL>} denotes using translation instruction written in natural language.} \label{tab:delimiter_results}
    \end{minipage}
\end{table}

%% file: sections/004-discussion.tex
\newpage
\section{Empirical Insights }\label{s:discussion}

\subsection{Monolingual Data Shapes Core Language Capabilities}\label{sec:mono}
 
LLMs have fundamentally transformed machine translation, primarily through their advanced capabilities derived from massive monolingual data. While previous studies~\cite{alma,tower_llm_2024} have demonstrated MT capabilities through direct finetuning of open-source LLMs, the specific impact of pretraining remains unclear. In this section, we perform preliminary experiments on ZH$\Leftrightarrow$EN translation with a 1.3B model. We lead the model to generate explanations together with translations to see how pretraining on large-scale monolingual data enhances LLM's understanding of the given sentence and further improves translation quality.

While monolingual data's primary contribution to translation tasks is often assumed to be improved fluency in generation, we also investigate several additional potential hypotheses as below:

\begin{itemize}
    \item[\rlap{\raisebox{0.3ex}{\hspace{0.4ex}\ding{56}}}$\square$] \textbf{Unlocking Natural Language Instruction Understanding}. Results show that models trained solely on parallel data can effectively follow diverse instructions and generate explanations after translation-oriented fine-tuning. This demonstrates that \textbf{the ability to follow natural language instructions is not exclusively acquired from monolingual data}.
    \item[\rlap{\raisebox{0.3ex}{\hspace{0.4ex}\ding{52}}}$\square$] \textbf{Enriching Factual Knowledge Base}. We ask annotators to evaluate the model's factual accuracy and semantic understanding based on its generated explanations. Results show that 200B monolingual data improves the factual accuracy from 59.1\% to 67.7\%. Table~\ref{tab:en_zh_example} shows that ``model w/ Mono.'' is able to handle complex contexts with typos (``\textbf{feveryone}'' in the first case) and translate the terminology correctly with additional knowledge (``\textbf{Lark 4.1}'' in the second case). This improvement suggests that large-scale monolingual data significantly enhances LLMs' ability to comprehend source sentences.
    \item[\rlap{\raisebox{0.3ex}{\hspace{0.4ex}\ding{52}}}$\square$] \textbf{Improving Reasoning Capabilities}. Results in Table~\ref{tab:mono_data} show that even the ``model w/ Mono.'' unexpectedly benefits from CoT patterns. However, the reasoning ability derived from parallel data remains limited, primarily due to the absence of code and mathematical data. 
\end{itemize}

\input{tables/mono_data_results}
\input{tables/en_zh_examples}

It is noted that parallel data also develops the above abilities that contributes to translation, but acquiring large-scale parallel data remains challenging. Therefore, a practical alternative is to leverage large amounts of accessible monolingual data for pretraining and then transfer the acquired knowledge to MT tasks.

\subsection{Parallel Data Quality and Usage Matter}

Prior research has proposed various ``optimal'' approaches for leveraging bilingual data in LLM-based machine translation, but our experiments demonstrate their limitations. \citeauthor{alma} cautioned against using excessive parallel data during LLM fine-tuning to avoid washing out the pre-existing knowledge. \citeauthor{tower_llm_2024} suggested that simply combining monolingual and parallel data would maximize performance. However, our work presents an alternative training strategy using parallel data and the key points are three-folds:

\begin{itemize}
    \item \textbf{Simple word-mapping sentence pairs are detrimental to translation quality}. Current open-source bilingual datasets suffer from significant quality issues, notably domain limitations, inadequate semantic alignment, and unnatural expressions. These deficiencies may have influenced the findings reported by \citeauthor{alma}. To address this, we spend much efforts in iterative filtering and refinement of the collected parallel data (detailed in Section~\ref{sec:parallel_data}), which is empirically validated in Table~\ref{fig:ablation_parallel}.
    \item \textbf{Continue training on pure parallel data maximizes translation effectiveness}. While this approach is generally avoided due to the risk of catastrophic forgetting, our experiments demonstrate its effectiveness when combined with high-quality parallel data and optimized learning rate scheduling. Furthermore, our investigation (Section~\ref{sec:transfer}) reveals intriguing insights into how parallel data influences multilingual transfer capabilities throughout this phase.
    \item \textbf{Over-training on pure parallel data may lead to overfitting}. Given the distributional shift during this process, we carefully monitor the trend of translation performance and stop training in time to prevent overfitting. While such degradation is often imperceptible through automated metrics, it becomes readily apparent during human evaluation.
\end{itemize}

\subsection{Translation Entails Thinking}

In existing open-source translation-oriented LLMs~\citep{tower_llm_2024,lu2024llamax}, the application of CoT reasoning to translation tasks remains largely unexplored. We propose that the model should systematically consider three key aspects: comprehending the source text's context, mastering domain-specific terminology and vocabulary, and formulating idiomatic expressions in the target language. We ask linguists to annotate the CoT as the above process, with an example illustrated in Table~\ref {tab:COT_case}. The phrase ``\begin{CJK*}{UTF8}{gbsn}做斗争\end{CJK*}'' in this context significantly deviates from its literal meaning, thus requiring an appropriate adjustment in translation. Results presented in Table~\ref {tab:sft_effect} demonstrate that CoT significantly contributes to improving translation accuracy.
\begin{CJK*}{UTF8}{gbsn}
\begin{table*}[t]
    \centering
    \small
    \begin{tabular}{p{2cm}p{13cm}}
    \toprule
    \textbf{Prompt} & Translate the following Chinese sentence into English and explain it in detail: 每次化妆都在跟自己极不对称的脸做斗争，今天拍视频的截图给大家对比下。 \\ \midrule
    \multirow{2}{*}{\textbf{CoT}} & \textbf{(Summary the meaning of the given sentence)} This is the description of having to deal with facial asymmetry every time the speaker puts on makeup, and intending to let everyone see the comparison by sharing screenshots of the video she took. \\
    & \textbf{(Explain unclear or difficult words)} ``做斗争'' refers to the speaker uses makeup techniques to adjust asymmetrical face. ``做斗争'' is a metaphor. It figuratively expresses that in the process of make-up, the make-up artist needs to put in efforts to deal with the asymmetry of the face in order to achieve a relatively satisfactory make-up effect. \\
    & \textbf{(Give the proper translation)} ``做斗争'' in this context does not convey a literal struggle or confrontation, but rather refers to ``making efforts to deal with'' something. Therefore, it should be translated as ``try to use makeup techniques to adjust asymmetrical face''.\\ \midrule
    \textbf{Translation} & Every time I put on makeup, I'm trying to use makeup techniques to adjust my asymmetrical face. Today, I've taken screenshots from a video to show you all a comparison. \\
    \bottomrule
    \end{tabular}
    \caption{A Chinese-to-English translation case study with annotated Chain-of-Thought process.}
    \label{tab:COT_case}
\end{table*}
\end{CJK*}

\newpage
\subsection{Mono-Parallel Balance Promotes Language Transfer} \label{sec:transfer}

\subsubsection{Progressive Training Strategy with Language Prioritization}

Prior work~\citep{zhao2024large} identifies the existence of a core language mechanism, whereby the model comprehends queries by converting multilingual inputs into English for task processing. This finding motivates our strategic organization of data ratios based on the distinction between core and side languages. We regard English as the primary language, which should constitute the dominant portion of the LLM's knowledge storage. As described in section~\ref{sec:training_Stage}, we continue with multilingual-dominant and parallel-only stages. Compared with direct mixing, this strategy can prevent the secondary languages from being overlooked due to data scarcity.

\subsubsection{Knowledge Transfer from Primary to Secondary Languages} 

\input{figures/tex/lm_ability}

This section investigates how generic knowledge can be leveraged to enhance translation capabilities, and examines the transfer of knowledge from primary to secondary languages under our proposed training strategy. We analyze the model performance starting from the completion of general pretraining through subsequent stages where translation capabilities are progressively enhanced through increased exposure to parallel data as \textbf{Stage I} and \textbf{Stage II}. We evaluate three fundamental capabilities: knowledge retention, language understanding, and reasoning ability.\footnote{We refer to OpenCompass~\citep{2023opencompass} to select datasets. For the task of knowledge examination, we choose ARC-easy/challenge and their multilingual version. For understanding tasks, we evaluate the model on OpenBookQA, TNEWS, AFQMC, C3, and iFLYTEK to report average accuracy for English and Chinese. For reasoning tasks, we select HellaSwag/COPA for English and CMNLI for Chinese.} Based on the observations in Table~\ref{fig:lm_ability}, we can conclude that:

\begin{itemize}
    \item \textbf{Parallel corpora allows knowledge transfer from English-similar to English-distant languages}. As shown in Figure~\ref{fig:lm_ability_exam}, English-distant languages demonstrate consistently improving accuracy in knowledge examination with the number of tokens consumed. Importantly, since our parallel data consists of multiple revisions of the original content, no new factual knowledge is introduced in later training stages. Thus, we attribute this growth to knowledge transfer from core languages.
    \item \textbf{Continue training on pure parallel data significantly enhances LLMs' understanding by establishing cross-lingual semantic equivalence}. As seen in Figure~\ref{fig:lm_ability_understand}, the performance of Chinese shows modest improvement during stage I but exhibits substantial gains in stage II. These significant improvements can be primarily attributed to the predominant proportion of Chinese-English parallel data in Stage II.
    \item \textbf{Certain performance trade-offs remain inevitable}. First, core language capabilities experience varying degrees of degradation during both two stages, attributed to the incorporation of additional languages. Second, English reasoning performance deteriorates significantly in stage II, primarily because the segmented nature of parallel corpora disrupts the coherence of monolingual long-form texts. With a limited model size, we need to sacrifice some capabilities that have a relatively small impact on translation. In the future, we will explore how to mitigate these losses under distribution shifts during continual training.
\end{itemize}

\begin{table*}[t]
    \centering
    \small
    \begin{tabular}{ccc}
    \toprule
    \textbf{Multi-parallel} & \multicolumn{2}{c}{\textbf{BLEURT}}\\ \cmidrule(lr){2-3}
    \textbf{Data} & \textbf{EN$\Rightarrow$XX} & \textbf{XX$\Rightarrow$EN} \\ 
    \midrule
    \ding{51}      & 71.63 & 75.44 \\
    \ding{55}      & \textbf{79.46} & \textbf{77.77} \\
    \bottomrule
    \end{tabular}
    \caption{Model performance comparison with and without multi-parallel data during instruction tuning.}
    \label{tab:sft_multi_para}
\end{table*}

\subsection{Overfitting Risks in Multilingual Instruction Data}\label{sec:sft}
Acquiring high-quality supervised multilingual translation data that covers a wide range of language directions remains a significant challenge. Although combining multi-parallel corpora seems like a straightforward solution to expand directional coverage, our empirical results, as shown in Table \ref {tab:sft_multi_para}, indicate that such data substantially impairs model performance during instruction fine-tuning. We attribute this phenomenon to the inherent overfitting risk posed by multi-parallel data's simplified learning patterns.
Based on these findings, we deliberately avoid indiscriminately increasing direction coverage through multi-parallel data, instead leveraging the LLM's generalization capability to handle uncovered directions.

%% file: tables/mono_data_results.tex
\begin{table*}[tb]
    \centering
    \small
    \begin{tabular}{ccccc}
    \toprule
    \multirow{2}{*}{\textbf{Mono.}} & \multirow{2}{*}{\textbf{CoT}}  & \multicolumn2{c}{\textbf{BLEURT}} & \multirow{2}{*}{\textbf{Factual Acc.}}\\ \cmidrule(lr){3-4}
    & & \textbf{ZH$\Rightarrow$EN} & \textbf{EN$\Rightarrow$ZH} & \\ \midrule
    \ding{55} & \ding{55} & 62.22 & 64.00 & - \\
    \ding{55} & \checkmark & 63.22 & 64.70 & 59.1\\
    \checkmark & \ding{55} & 63.55 & 65.88 & -\\
    \checkmark &  \checkmark & \textbf{64.59} &  \textbf{66.62} & \textbf{67.7}\\
    \bottomrule
    \end{tabular}
    \caption{Translation quality on in-house ZH$\Leftrightarrow$EN testsets with different settings. ``w/o Mono.'' means the model is purely trained on concatenated parallel data during pretraining, ``CoT'' stands for the model is trained with a chain of task: translate-then-explain. We ask annotators to check whether the model makes factual errors or misunderstands the semantics of the given sentence based on generated explanations as factual accuracy.}
    \label{tab:mono_data}
\end{table*}

%% file: tables/en_zh_examples.tex
\begin{CJK*}{UTF8}{gbsn}
\begin{table*}[t]
    \centering
    \small
    \begin{tabular}{p{4cm}p{5cm}p{5cm}}
    \toprule
    \textbf{Prompt} & \textbf{Model w/o Mono.} & \textbf{Model w/ Mono.} \\ \midrule
    Translate the sentence to Chinese and try to explain: Thank you \textbf{feveryone} for helping us grow spiritually! & 感谢 \textcolor{red}{\textbf{Youfeone}} 帮助我们在精神上成长！(This sentence is a polite way to thank \textcolor{red}{\textbf{someone named Youfeone}} for helping the speaker grow spiritually.) \ding{55} & 感谢\textcolor{teal}{\textbf{你们}}帮助我们在精神上成长！(This sentence is a thank you message to someone who has helped the speaker to grow spiritually.) \checkmark\\ \midrule
    Translate the sentence to Chinese and explain it: From \textbf{Lark 4.1} onwards, you can now add up to 10 group chat administrators. & 从\textcolor{red}{\textbf{4.1}}开始，您现在可以添加最多10名群聊管理员。(This sentence is related to adding group chat administrators to a group chat. \textcolor{red}{\textbf{It starts with the date "4.1"}} and describes the action that can be done from that date.) \ding{55} & 从\textcolor{teal}{\textbf{Lark 4.1}}开始，您现在可以添加最多10个群聊管理员。(\textcolor{teal}{\textbf{Lark is a popular messaging and collaboration app in China. ``4.1'' refers to the release of version 4.1.}} ``group chat administrators'' are administrators who manage and configure group chats within a group.) \checkmark\\
    \bottomrule
    \end{tabular}
    \caption{EN$\Rightarrow$ZH translation examples. The \textbf{bold} words in the prompt are translated in different ways by these two models. The colored parts are the corresponding translations and explanations. \ding{55} denotes the incorrect interpretation of the input sentence judged by annotators and vice versa.}\label{tab:en_zh_example}
\end{table*}
\end{CJK*}

%% file: figures/tex/lm_ability.tex
\begin{figure*}[t]
     \centering
     \begin{subfigure}[b]{0.32\textwidth}
         \centering
         \includegraphics[width=\textwidth]{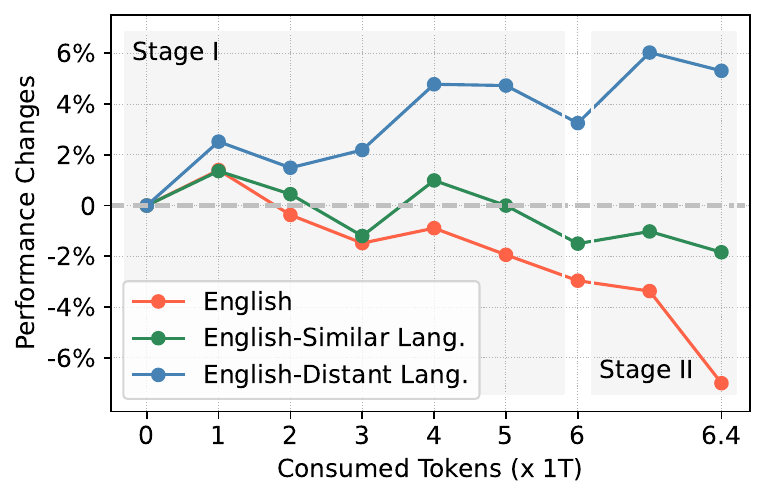}
         \caption{Knowledge Examination}
         \label{fig:lm_ability_exam}
     \end{subfigure}
     \hfill
     \begin{subfigure}[b]{0.32\textwidth}
         \centering
         \includegraphics[width=\textwidth]{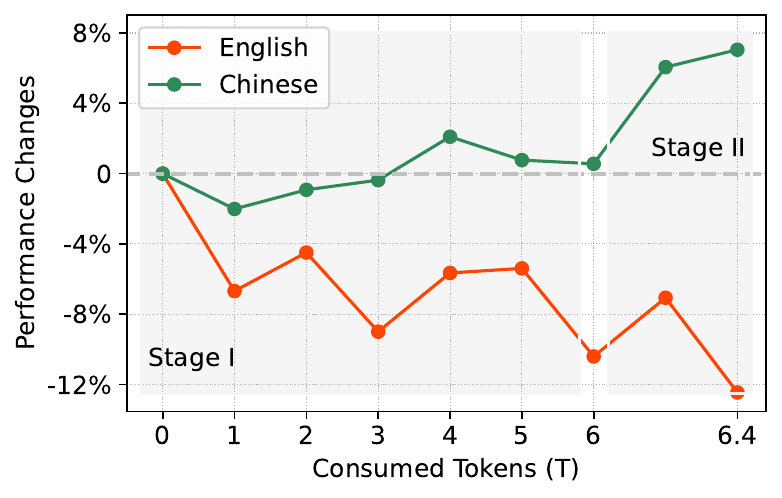}
         \caption{Understanding Tasks}
         \label{fig:lm_ability_understand}
     \end{subfigure}
     \hfill
     \begin{subfigure}[b]{0.32\textwidth}
         \centering
         \includegraphics[width=\textwidth]{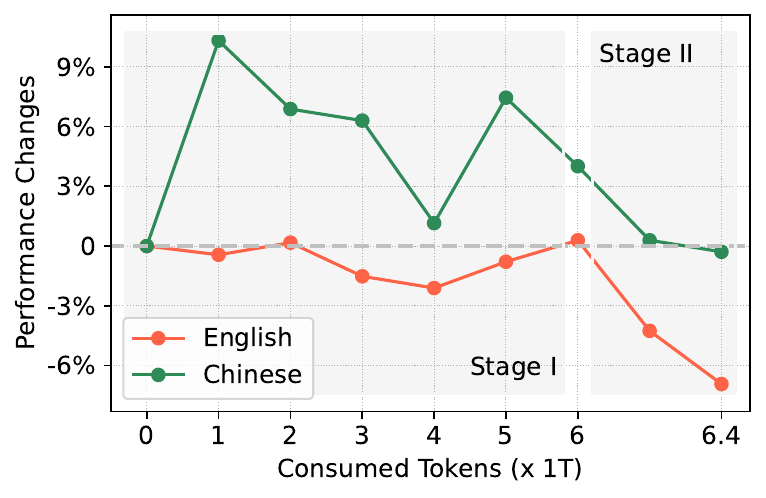}
         \caption{Reasoning Tasks}
         \label{fig:lm_ability_reason}
     \end{subfigure}
        \caption{Performance changes on benchmarks in different stages of pretraining (\textbf{Stage I}: mixed corpora with an increased ratio of parallel data, \textbf{Stage II}: parallel data only). English-distant languages include Japanese, Thai, Russian, Malay, Indonesian, and Arabic, while English-similar languages refer to German, Spanish, Portuguese, and French.}\label{fig:lm_ability}
\end{figure*}

%% file: sections/005-related.tex
\section{Related Work}\label{s:related_work}

Recent work has explored beating ultra-large models in specific abilities by pushing small-scale models with maximum leverage. In this paper, we extend the limitations of the 7B model as a professional interpreter recognized by human annotators. This section reviews this trend of tailoring LLMs for specific tasks, especially in the field of machine translation.

\textbf{Specialization of LLMs.} The specialization aims at higher accuracy and more expert-level responses in the respective fields. Specialized LLMs are trained on targeted datasets that are relevant to a particular domain, such as law~\cite{li2023sailer,cui2023chatlaw} and medicine~\cite{zhou2023survey}, or specific tasks like mathematical reasoning~\cite{shao2024deepseekmath,wang2023mathcoder} and code~\cite{roziere2023code,li2023starcoder,luowizardcoder}. These models can be developed in two ways: (1) pre-training from scratch with a combination of general and domain-related text, and (2) starting from an open source model and adapting to the target field through continuous training and task-specific supervised fine-tuning~\cite{yue2023mammoth}. Although the second method is significantly cheaper, the overall performance may be somewhat limited by the initial capabilities of the base model. Thus, we train from scratch to push the limits of multilingualism in pre-training to lay a solid foundation for the translation task.

\textbf{Translation-oriented LLMs.} Much effort has been made to match the cutting-edge multilingual translation capabilities of GPT-4. The TowerInstruct 13B model~\citep{tower_llm_2024} outperforms open alternatives but lags behind GPT-4 in automatic evaluation. The ALMA series models~\citep{alma,almar} perform better than the NLLB-54B model~\citep{nllb} and GPT-3.5, but still fall short of GPT-4. Importantly, existing research only reports automatic evaluation on public MT datasets, despite ultra-large models often outperforming others when evaluated by humans on more varied and authentic translations, especially in challenging scenarios~\citep{jiao2023chatgpt}. This paper makes remarkable progress, surpassing ultra-large models on both automatic metrics and human evaluation. We also release a challenging MT benchmark to set a higher standard for conciseness, accuracy, and elegance.

Balancing multilingual capabilities has been crucial in machine translation~\citep{li2024eliciting}, especially for low-resource languages~\citep{zou2025trans}. This paper proposes an efficient training approach to maximize multilingual translation ability and explores a new perspective on cross-lingual knowledge transfer from primary to secondary languages.

%% file: sections/006-conclusion.tex
\section{Conclusion}
In this paper, we present \method, a family of open-source LLMs designed specifically for machine translation, supporting bidirectional translation across 28 languages. We provide a detailed introduction of the complete training pipeline, including pre-training, instruction tuning, and reinforcement learning, while sharing key insights and best practices derived from our hill-climbing optimization process. Remarkably, with only 7B parameters, \method achieves translation quality comparable to, or even surpassing, the most advanced LLMs and leading commercial translation systems, as evidenced by both automatic and human evaluations. By releasing \method's publicly available weights, we aim to make \method an accessible, off-the-shelf tool to support the community in advancing translation research and applications.

%% file: appendix/language.tex
\section{Supported Languages}\label{s:support_languages}
\begin{table*}[ht]
\centering
\small
\begin{tabular}{cccc}
\toprule
\textbf{Languages}  & \textbf{Abbr.} & \textbf{Languages}        & \textbf{Abbr.} \\ \midrule
Arabic              & ar              & Malay                     & ms     \\
Czech               & cs                    & Norwegian Bokmal          & nb                    \\
Danish              & da                    & Dutch                     & nl                    \\
German              & de                    & Norwegian                 & no                    \\
English             & en                    & Polish                    & pl                    \\
Spanish             & es                    & Portuguese                & pt                    \\
Finnish             & fi                    & Romanian                  & ro                    \\
French              & fr                    & Russian                   & ru                                \\
Croatian            & hr                    & Swedish                   & sv                    \\
Hungarian           & hu                    &  Thai                      & th      \\
Indonesian          & id                    & Turkish                   & tr                    \\
Italian             & it                    & Ukrainian                 & uk           \\
Japanese            & ja                    & Vietnamese                & vi                    \\
Korean              & ko                    & Chinese                   & zh                    \\
\bottomrule
\end{tabular}
\caption{Supported languages.}
\label{tab:test_langs}
\end{table*}

%% file: appendix/experimental_set.tex
\section{Implementation Details}\label{s:Experimental_Settings}
\textbf{Architecture}. To enhance accessibility and adoption within the open-source community, we construct our 7B model following the architecture of Mistral-7B~\citep{jiang2023mistral}. This architecture is based on the Transformer~\citep{vaswani2017attention}, and it consists of 32 Transformer decoder layers, each containing 32 self-attention heads, an embedding size of 4,096 dimensions, and a feed-forward neural network size of 14,336 dimensions. Layer normalization is used to stabilize learning. Additionally, to improve the model's ability to handle positional information, the model equipped with Rotary Position Embedding~\citep[ROPE,][]{rope}. ROPE modifies the standard positional encoding by using a rotation matrix, allowing the model to capture relative positions more effectively. Additionally, we set the maximum sequence length to 2,048, precisely calibrated to match our real-world translation requirements.

\textbf{Vocabulary}. Building upon established techniques, we design our tokenization approach utilizing the Byte Pair Encoding~\citep[BPE,][]{bpe} method, similar to what Mistral uses, with a significant enhancement by expanding the original vocabulary from 32,000 to 65,269 tokens. Compared to the original vocabulary, our enhanced tokenizer boosts compression rates from 3.17 to 3.74 characters per token across a diverse set of multilingual data.
This strategic extension enables our model to better handle the multilingual data and nuances of translation tasks, ensuring more comprehensive linguistic coverage. 

\textbf{Training}. Through rigorous experimentation, we methodically test and optimize various batch sizes and learning rates. We find that utilizing a batch size of 2M tokens alongside a learning rate of 3e-4 constitutes the optimal setup for pretraining. The models undergo a warm-up phase involving 2,000 linear scaling steps to reach the peak learning rate, followed by cosine decay to gradually reduce it to a minimum of 10\% of the peak rate. During the supervised fine-tuning stage, we utilize a batch size of 64 sentences and a learning rate of 3e-6, striking a precise balance that refines the model’s performance, ultimately leading to superior translation quality.

\textbf{Inference}. During inference, we utilize beam search (with a beam size of 4) to generate translation results. We observed that beam search slightly outperforms greedy decoding, though at the expense of inference speed.

\begin{figure}[!h]
    \centering
    \includegraphics[width=1\linewidth]{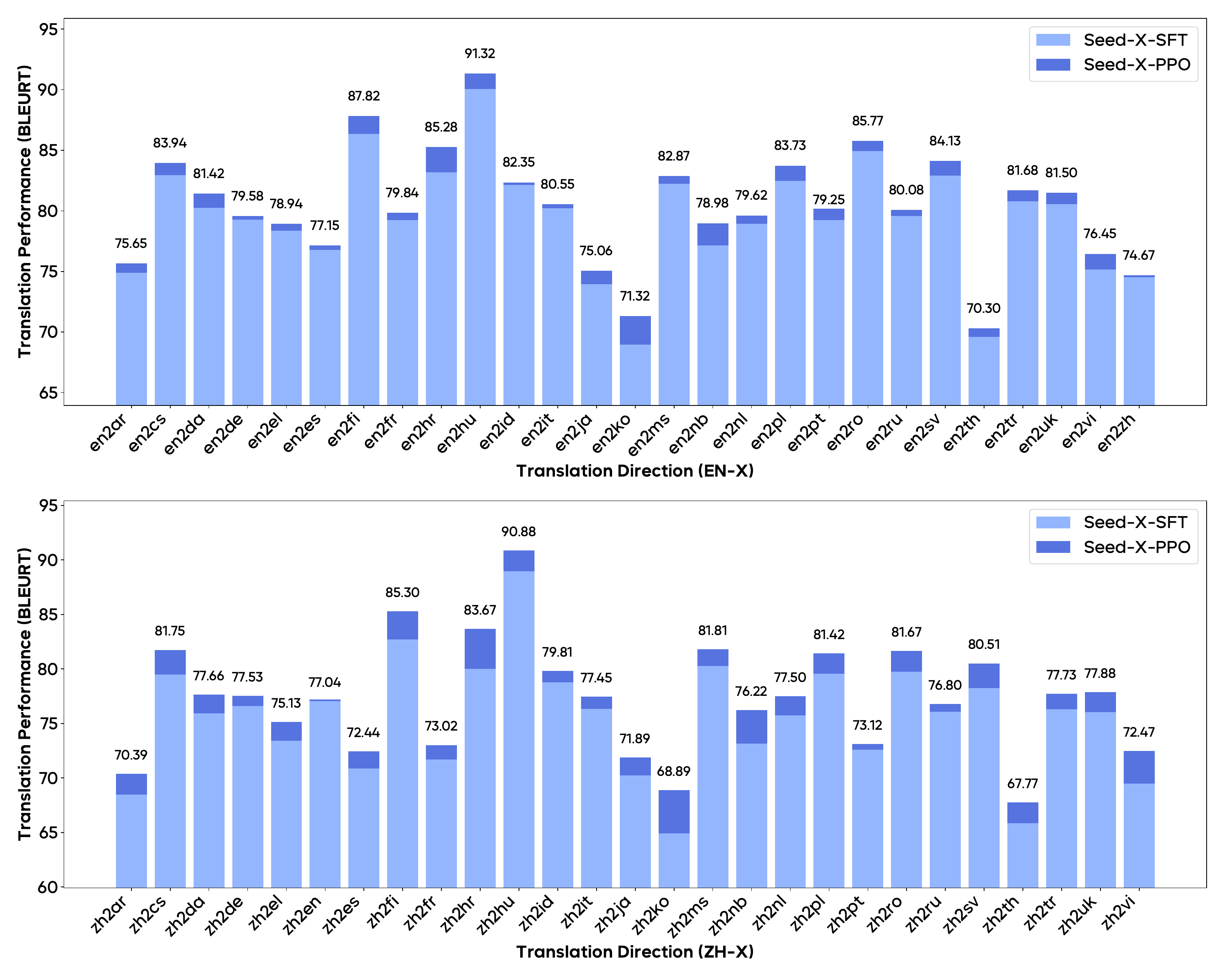}
    \caption{Performance comparison between Seed-X-SFT and Seed-X-PPO across evaluated translation directions on Flores-200.}
    \label{fig:diff_sft_ppo}
\end{figure}

%% file: appendix/ppo_vs_sft.tex
\section{Translation Performance Across Individual post-training stages}\label{s:ppo_vs_sft}
In Figure \ref{fig:diff_sft_ppo}, we present the the detailed performance of \method across multiple translation directions after instruction-tuning (SFT) and reinforcement learning (PPO).
The results show that, after the SFT stage, the model already achieves strong translation quality, as measured by BLEURT scores.
PPO further improves upon the SFT model, pushing the performance ceiling for machine translation.

%% file: appendix/human_eval.tex
\section{Human Evaluation} \label{appendix:human_evaluation}

\begin{figure}
    \centering
    \includegraphics[width=0.45\linewidth]{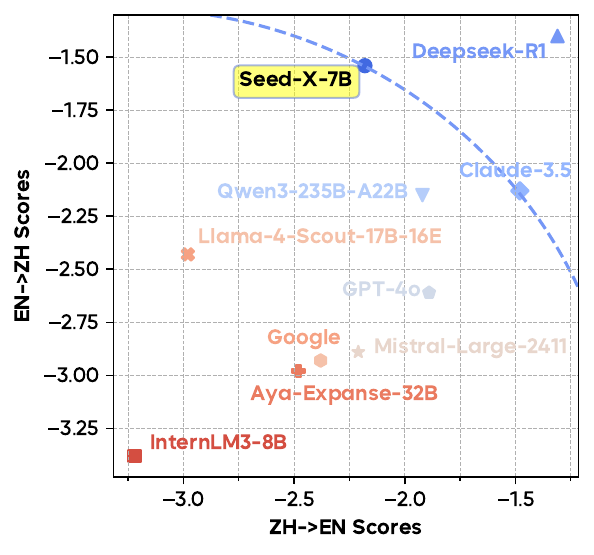}
    \caption{Human evaluation of models on the \textbf{\method-Challenge} for the Chinese-English translation task. We use a strict deduction system to evaluate the quality of translations, accumulating deductions for major and minor errors in the translated text, so the final scores presented are all negative numbers. The average Chinese-English translation ability of the models within the arc in the figure is worse than that of \method.}
    \label{fig:human_eval_zhen}
\end{figure}

Figure~\ref{fig:human_eval_zhen} shows the human evaluation results for the Chinese-English translation task. For this high-resource language pair, we use a strict deduction system to evaluate translation quality, accumulating deductions for major and minor errors in the translated text. It demonstrates that \method outperforms most models with a distinct advantage on EN$\Rightarrow$ZH, trailing only behind Deepseek-R1. Table~\ref{tab:detailed_human_scores} shows the human evaluation scores for each translation direction corresponding to Figure~\ref{fig:humen_eval}. In 14 language directions, \method-PPO achieved 6 first places and one second place, with the highest overall average score, showing a clear advantage in EN$\Rightarrow$XX language directions.

\begin{table*}[!t]
\centering
\small
\tabcolsep 3pt
\begin{tabular}{p{4.3cm}cccccccccccccccc}
\toprule
 \multirow{2}{*}{\textbf{Models}}  & \multicolumn{7}{c}{\textbf{EN$\Rightarrow$XX}} &  \multicolumn{7}{c}{\textbf{XX$\Rightarrow$EN}} & \multirow{2}{*}{\textbf{Avg.}} \\ \cmidrule(lr){2-8} \cmidrule(lr){9-15}
       & \textbf{es}   & \textbf{de}   & \textbf{fr}   & \textbf{ru}   & \textbf{pt}  & \textbf{it}   & \textbf{ar}   & \textbf{es}   & \textbf{de}   & \textbf{fr}   & \textbf{ru}   & \textbf{pt}  & \textbf{it}   & \textbf{ar}  \\ \midrule
InternLM3-8B$^\dagger$~\citep{cai2024internlm2}& 2.56 &2.77 &3.11 &2.70 &3.74 &2.60 &1.69 &3.35 &1.85 &3.06 &3.16 &3.07 &3.05 &2.50 & 2.80 \\
LLaMAX3-8B$^\dagger$~\citep{lu2024llamax} & 2.85 &3.04 &3.30 &3.16 &3.76 &3.11 &2.66 &3.23 &2.10 &3.06 &3.55 &2.72 &3.10 &3.26 & 3.06 \\
Google-Translator & 3.66 &3.55 &3.44 & \uline{3.77} &3.84 &3.42 &3.53 &3.29 &3.32 &2.99 &3.50 &3.11 &3.19 &3.57 & 3.44 \\
Mistral-Large-2411-123B$^\dagger$~\citep{Mistral}& 3.55 &3.61 &3.38 &3.64 &3.79 &3.31 &3.55 &3.30 &3.20 &3.28 &3.68 &3.07 &3.53 &3.60 & 3.46 \\
Llama-4-Scout-17B-16E$^\dagger$~\citep{meta2025llama}& 3.62 &3.60 &3.62 &3.66 &3.86 &3.50 &3.38 &3.56 &3.18 &3.70 &3.47 &3.31 &3.65 &3.60 & 3.55 \\
Gemma3-27B$^\dagger$~\citep{gemma}& 3.72 & \textbf{3.66} &3.43 &3.76 &3.81 &3.39 &3.57 &3.75 &3.47 &3.50 &3.75 &3.29 &3.61 &3.73 & 3.60 \\
Qwen3-235B-A22B$^\dagger$~\citep{yang2025qwen3}& 3.62 &3.56 &3.75 &3.69 &3.87 &\uline{3.61} &3.25 &3.56 &3.32 &\uline{3.77} &3.76 &3.41 &3.74 &3.64 & 3.61 \\
Aya-32B$^\dagger$~\citep{dang2024aya}& 3.59 &3.58 & \uline{3.79} &3.75 &3.88 & \textbf{3.64} &3.60 &3.73 &3.25 &3.66 &3.25 &3.34 &3.72 &3.89 & 3.62 \\
DeepSeek-V3$^\dagger$~\citep{liu2024deepseek}& 3.81 &3.53 &3.60 &3.73 &3.81 &3.31 &3.53 &3.80 &3.52 &3.69 &3.79 &3.37 &3.55 &\uline{3.91} & 3.64 \\
DeepSeek-R1$^\dagger$~\citep{deepseekai2025deepseekr1}& 3.82 &3.57 &3.51 &3.69 &3.81 &3.31 &3.59 &3.75 &\uline{3.66} &\textbf{3.79} & \textbf{3.81} &3.21 &3.64 &3.84 & 3.64 \\
GPT-4o~\citep{GPT-4o} & 3.74 & \uline{3.65} &3.63 &3.74 &3.82 &3.60 &3.54 &3.58 &3.27 &3.62 &3.63 &\textbf{3.67} &3.84 & \textbf{3.91} & 3.66 \\
Gemini-2.5-Pro~\citep{gemini25} & 3.76 &3.59 &3.53 &3.61 & \uline{3.88} &3.58 &\uline{3.67} &3.69 &3.42 &3.85 &3.52 &\uline{3.63} &\uline{3.86} &3.68 & 3.66 \\
Claude-3.5-Sonnet~\citep{claude35} & \uline{3.84} &3.50 &3.55 &3.75 &3.82 &3.48 &3.59 &\textbf{3.87} & \textbf{3.76} &3.67 &\uline{3.79} &3.35 &3.69 &3.87 & \uline{3.68} \\ \midrule
\textbf{\method-PPO} & \textbf{3.87} &3.61 & \textbf{3.88} & \textbf{3.88} & \textbf{3.88} &3.55 & \textbf{3.78} & \uline{3.85} &3.49 &3.74 &3.55 &3.55 & \textbf{3.87} &3.75 & \textbf{3.73} \\ \bottomrule
\end{tabular}
\caption{Human evaluation scores (0-4) on the \textbf{\method-Challenge} for 14 language directions corresponding to Figure~\ref{fig:humen_eval}. Open-source models are marked with $\dagger$. Models are arranged in ascending order by average score, with \textbf{the highest score} in bold and \uline{the second-highest score} underlined.} \label{tab:detailed_human_scores}
\end{table*}

%% file: appendix/case_study_new.tex
\section{Case Study}\label{s:case_study}
Table \ref{tab:case_study} presents representative Chinese-English translation examples, highlighting key testing points and corresponding translations by \method. These cases demonstrate \method's contextual awareness and competence in handling complex translations.
\newpage
\begin{CJK*}{UTF8}{gbsn}
\begin{longtable}{p{2.4cm}p{12.5cm}}
    \toprule
    \multicolumn{2}{c}{\textbf{Scenario 1: Internet buzzwords}} \\
    \midrule
    \textbf{Input \#1} & 在某宝上看到一件裙子\ 绝绝子\ 红色YYDS，加油瘦\ 然后去拥有它！\\
    \textbf{Testing Points} & ``某宝'': Taobao, ``绝绝子'': amazing/incredible, ``YYDS'': forever the best, ``加油瘦'': the subject is the speaker. \\
    \textbf{Google} & I saw a skirt on Taobao. It’s \textcolor{red}{\textbf{absolutely red. YYDS}}, go ahead and lose weight, then \textcolor{red}{\textbf{go get it}}. \\
    \textbf{\method} & I saw a skirt on Taobao that's absolutely amazing! Red is \textcolor{teal}{\textbf{the ultimate choice}}! Keep working hard to \textcolor{teal}{\textbf{lose weight and then go get it'll be yours}}!\\
    \midrule
    \textbf{Input \#2} & 光腿神器\\
    \textbf{Testing Points} & ``光腿神器'': a pair of leggings that look just like bare legs when worn. \\
    \textbf{Google} & \textcolor{red}{\textbf{Bare Legs Artifact}} \\
    \textbf{\method} & \textcolor{teal}{\textbf{Pantyhose that make legs look bare}}\\

    \midrule
    \textbf{Input \#3} & INTJ总是装E\\
    \textbf{Testing Points} & ``E'': in psychology, it usually refers to people with an extraverted personality. \\
    \textbf{Google} & INTJ always pretends to be \textcolor{red}{\textbf{E}}? \\
    \textbf{\method} & INTJs always pretend to be \textcolor{teal}{\textbf{extraverts}}?\\

    \midrule
    \textbf{Input \#4} & 集美们在集美区做什么？\\
    \textbf{Testing Points} & ``集美们'': it has the similar pronunciation with ``姐妹'' (girls). \\
    \textbf{Google} & What do \textcolor{red}{\textbf{Jimei}} people do in Jimei District?\\
    \textbf{\method} & What are you \textcolor{teal}{\textbf{girls}} doing in Jimei District?\\

    \midrule
    \textbf{Input \#4} & u1s1，这次城市峰会真是泰裤辣！\\
    \textbf{Testing Points} & ``u1s1'': it is an abbreviation, which means ``to be honest''. ``泰裤辣'': it has the same pronunciation with ``太酷啦'', which means ``awesome''.\\
    \textbf{Google} & \textcolor{red}{\textbf{u1s1}}, this city summit is really \textcolor{red}{\textbf{hot!}}\\
    \textbf{\method} & \textcolor{teal}{\textbf{To be honest}}, this city summit is really \textcolor{teal}{\textbf{awesome!}}\\
    \midrule
    
    \multicolumn{2}{c}{\textbf{Scenario 2: Chinese Slang}} \\
    
    \midrule
    \textbf{Input \#6} & 不撞南墙是会不知道有多痛的。。。疼了痛了才会明白！\\
    \textbf{Testing Points} & ``撞南墙'': to learn something the hard way. \\
    \textbf{Google} & You won’t know how painful it is until you \textcolor{red}{\textbf{hit your head against the wall}}... You won’t understand until you feel the pain! \\
    \textbf{\method} & You won't know how painful it is until you \textcolor{teal}{\textbf{hit a wall head-on}}... Only when it hurts will you realize it! \\
    \midrule
    \textbf{Input \#5} & 有点厉害，不知道水分大不大\\
    \textbf{Testing Points} & ``水分'': data or descriptions that contain exaggerated or untrue elements. \\
    \textbf{Google} & It's a bit powerful, I don't know if it has a lot of \textcolor{red}{\textbf{water}}.\\
    \textbf{\method} & That's pretty impressive. I wonder if there's \textcolor{teal}{\textbf{a lot of hype}} around it. \\
    \midrule
    \textbf{Input \#6} & 我说一句你说一车啊 \\
    \textbf{Testing Points} & ``说一车'': an exaggerated expression, which means that there's a lot to say.\\
    \textbf{Google} & I said one thing, you said \textcolor{red}{\textbf{one car}}\\
    \textbf{\method} & I say one thing, and you come up with \textcolor{teal}{\textbf{a whole bunch of responses}}\\
    \midrule
    \multicolumn{2}{c}{\textbf{Scenario 3: Classical Chinese}} \\
    \midrule
    \textbf{Input \#8} & 若夫淫雨霏霏，连月不开，阴风怒号，浊浪排空 \\ 
    \textbf{Testing Points}  & 释义：像那阴雨连绵，接连几个月不放晴，寒风怒吼，浑浊的浪冲向天空。 \\
    \textbf{Google} & If the rain is \textcolor{red}{\textbf{heavy and the moon is not bright}}, the wind is howling, and the turbid waves are rolling \\
    \textbf{\method} & As for the \textcolor{teal}{\textbf{persistent}} rain, it lasted for several months without stopping. The \textcolor{teal}{\textbf{gloomy wind howled furiously}}, and the turbulent waves filled the sky. \\
    \midrule
    \textbf{Input \#9} & 桃李不言，下自成蹊。 \\ 
    \textbf{Testing Points}  & 释义：桃树、李树不会说话，但因其果实可口，人们纷纷去摘取，于是便在树下踩出一条路来。比喻只要人品高尚，就能得到别人的尊敬和景仰。\\
    \textbf{Google} & The peaches and plums do not speak, but \textcolor{red}{\textbf{they create a trail of their own}}. \\
    \textbf{\method} & \textcolor{teal}{\textbf{A good example always speaks for itself, and others will naturally follow suit.}} \\
    \midrule
    \textbf{Input \#10} & 天高地迥，觉宇宙之无穷；兴尽悲来，识盈虚之有数。 \\ 
    \textbf{Testing Points}  & 释义：苍天高远，大地寥廓，令人感到宇宙的无穷无尽。欢乐逝去，悲哀袭来，意识到万事万物皆有定数。 \\
    \textbf{Google} & The sky is high and the earth is vast, and one feels the infinity of the universe; when the excitement ends and sadness comes, one realizes \textcolor{red}{\textbf{the finiteness of fullness and emptiness}}. \\
    \textbf{\method} & The sky is high and the earth is vast, making me feel the boundlessness of the universe. When my joy reaches its peak, sorrow inevitably follows, and I realize that \textcolor{teal}{\textbf{all things in the world have their own limits}}. \\
    \midrule

    \multicolumn{2}{c}{\textbf{Scenario 4: English Slang}} \\
    \midrule

    \textbf{Input \#12} & Their relationship is a total situationship. \\ 
    \textbf{Testing Points}  & It's the ``grey area'' where two people are involved romantically but haven't defined what they are to each other. \\
    \textbf{Google} & 他们的关系完全是\textcolor{red}{\textbf{一种关系}}。 \\
    \textbf{\method} & 他们的关系完全是\textcolor{teal}{\textbf{逢场作戏}}。\\
    \midrule

    \textbf{Input \#13} & I'm totally hangry right now. \\ 
    \textbf{Testing Points}  & ``hangry'': it combines the words ``hungry'' and ``angry'' to describe the bad mood. \\
    \textbf{Google} & \textcolor{red}{\textbf{我现在非常饿}}。 \\
    \textbf{\method} & \textcolor{teal}{\textbf{我现在真是又饿又生气}}。\\
    \midrule

    \textbf{Input \#14} & Out of the frying pan and into the fire. \\ 
    \textbf{Testing Points}  & It means you've escaped from one bad situation to end up in an even worse one. \\
    \textbf{Google} & \textcolor{red}{\textbf{刚从油锅里跳进火里}}。 \\
    \textbf{\method} & \textcolor{teal}{\textbf{真是屋漏偏逢连夜雨}}。\\    
    \bottomrule
    \caption{Comparative case studies across different scenarios. Red text highlights translation errors in Google Translate outputs, demonstrating \method's enhanced accuracy.}
    \label{tab:case_study}
\end{longtable}
\end{CJK*}

%% file: appendix/contributions.tex
\renewcommand{\thefootnote}{}
\section{Contributions}

\textbf{Project Lead}

Shanbo Cheng

\textbf{Core Contributors}

Yu Bao, Qian Cao\textsuperscript{1}\footnotetext{1 indicates the contributor has already left ByteDance. We appreciate their contribution to this work.}, Luyang Huang, Liyan Kang, Zhicheng Liu, Yu Lu, Wenhao Zhu

\textbf{Contributors}

Jingwen Chen\textsuperscript{1}, Zhichao Huang, Tao Li, Yifu Li, Huiying Lin\textsuperscript{1}, Sitong Liu, Ningxin Peng, Shuaijie She\textsuperscript{2}, Lu Xu, Nuo Xu, Sen Yang\textsuperscript{2}\footnotetext{2 indicates work done during internship, Nanjing University.}, Runsheng Yu, Yiming Yu, Liehao Zou

\textbf{Supervision}

Hang Li, Lu Lu, Yuxuan Wang, Yonghui Wu

(Last-Name in Alphabetical Order)